\documentclass[letterpaper, 10 pt, conference]{ieeeconf}  
\usepackage{graphicx}
\usepackage{gensymb}
\usepackage{amssymb}
\usepackage{color}

\IEEEoverridecommandlockouts                              

\title{\LARGE \bf
HPL-ViT: A Unified Perception Framework for Heterogeneous\\ Parallel LiDARs in V2V
}

\author{Yuhang Liu$^{1}$, Boyi Sun$^{2}$, Yuke Li$^{3}$, Yuzheng Hu$^{4}$, Fei-Yue Wang$^{5}$
\thanks{$^{1}$Yuhang Liu, and Fei-Yue Wang are with the Institute of Automation, Chinese Academy of Science, Beijing, China.
        {\tt\small liuyuhang2021@ia.ac.cn, feiyue.wang@ia.ac.cn}}%
\thanks{$^{2}$Boyi Sun is with the Department of Computer Science and Technology, University of Chinese Academy of Sciences, Beijing, China.
        {\tt\small sunboyi20@mails.ucas.ac.cn}}%
\thanks{$^{3}$Yuke Li is with Waytous Inc., Qingdao, China.
        {\tt\small liyuke14@mails.ucas.ac.cn}}%
\thanks{$^{4}$Yuzheng Hu is with the Department of Computer Science, University of Illinois, Urbana-Champaign, IL, USA.
        {\tt\small yh46@illinois.edu}}%
}

\begin{document}

\maketitle
\thispagestyle{empty}
\pagestyle{empty}

\begin{abstract}

To develop the next generation of intelligent LiDARs, we propose a novel framework of parallel LiDARs and construct a hardware prototype in our experimental platform, DAWN (Digital Artificial World for Natural).
It emphasizes the tight integration of physical and digital space in LiDAR systems, with networking being one of its supported core features. 
In the context of autonomous driving, V2V (Vehicle-to-Vehicle) technology enables efficient information sharing between different agents which significantly promotes the development of LiDAR networks. 
However, current research operates under an ideal situation where all vehicles are equipped with identical LiDAR, ignoring the diversity of LiDAR categories and operating frequencies. 
In this paper, we first utilize OpenCDA and RLS (Realistic LiDAR Simulation) to construct a novel heterogeneous LiDAR dataset named OPV2V-HPL. 
Additionally, we present HPL-ViT, a pioneering architecture designed for robust feature fusion in heterogeneous and dynamic scenarios. 
It uses a graph-attention Transformer to extract domain-specific features for each agent, coupled with a cross-attention mechanism for the final fusion. 
Extensive experiments on OPV2V-HPL demonstrate that HPL-ViT achieves SOTA (state-of-the-art) performance in all settings and exhibits outstanding generalization capabilities.

\end{abstract}

\section{INTRODUCTION}

LiDAR sensor plays a crucial role in vehicle perception systems that enables the understanding of 3D structural information.
Driven by the rapid development of artificial intelligence and communication technologies, LiDAR systems are steadily advancing toward digitization, networking, and intelligence \cite{9455394}.
To construct smart LiDARs, we propose a novel parallel LiDAR framework based on the theory of parallel intelligence \cite{liu2022parallel, liu2022radarverses}.
It aims to efficiently utilize software systems to enhance the sensing capabilities of physical LiDARs, emphasizing the significance of digital space in LiDAR systems.
We have also developed a hardware prototype using the parallel sensing platform known as DAWN \cite{10163892}.
Parallel LiDARs can utilize software to redefine hardware operations to achieve adaptive perception of the scene. 
The full name of DAWN is “Digital Artificial World for Natural” and it is a comprehensive platform designed for developing the next-generation intelligent sensors.
In this paper, the networking capabilities of parallel LiDARs with a focus on autonomous driving scenarios will be investigated.

\begin{figure}[!t]
\centering
\includegraphics[width=3.2in]{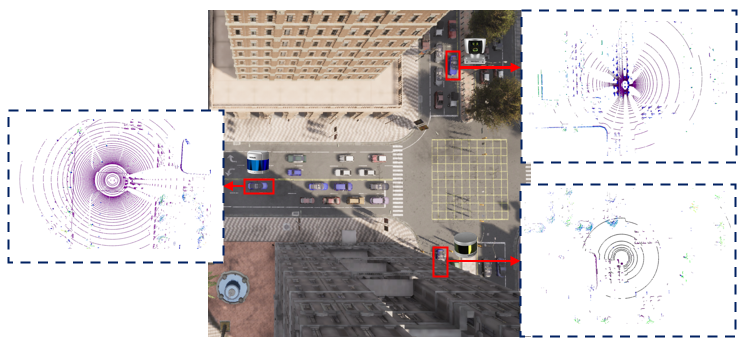}
\caption{A heterogeneous scenario with different LiDARs.}
\label{intro}
\end{figure}

V2V (Vehicle-to-Vehicle) technology enables efficient data sharing among autonomous vehicles which can effectively improve their perception performance \cite{9732063}. 
Compared to sharing raw data \cite{gao2018object} and detection results \cite{song2023cooperative, 10161460}, intermediate feature sharing is the prevailing method in V2V cooperative perception \cite{xu2022v2x, lei2022latency, hu2022where2comm, 10077757}.
It can not only achieve high perception accuracy but also conserve valuable communication bandwidth. 
Currently, several datasets have been released for evaluating fusion methods in V2V, such as OPV2V \cite{xu2022opv2v} collected in the CARLA simulator \cite{dosovitskiy2017carla}, and V2V4Real \cite{xu2023v2v4real} gathered from real-world scenarios. 
However, these datasets all assume that each agent is equipped with an identical LiDAR, representing a significant simplification of real scenes.
As illustrated in Figure \ref{intro}, it is common to observe that different LiDAR systems are employed in the same scenario. 
Besides, each vehicle also has the flexibility to adjust the working frequency of its LiDARs, introducing an additional heterogeneity of frequency. 
Furthermore, the proposed software-defined adaptive parallel LiDARs will exacerbate sensor heterogeneity in the real application.

In this work, our primary focus lies in achieving robust cooperative perception among heterogeneous parallel LiDARs in V2V, with the aim of minimizing model performance degradation in complex scenarios.
To investigate this issue, our first step involves the collection of a new dataset that encompasses diverse LiDAR sensors.
Considering the high costs of collecting data in physical contexts, we use OpenCDA \cite{9564825} in conjunction with RLS (Realistic LiDAR Simulation) \cite{10161027} to construct a novel dataset in the CARLA simulator known as OPV2V-HPL.  
We replay all scenes from OPV2V and perform data collection using four high-fidelity LiDAR models in RLS at two different operating frequencies. 
Then a novel HPL-ViT (\textbf{H}eterogeneous \textbf{P}arallel \textbf{L}iDARs-\textbf{Vi}sion \textbf{T}ransformer) framework for feature fusion among different LiDARs is introduced.
It effectively leverages prior information regarding LiDAR categories and operating frequencies to optimize its performance.
Each vehicle initially generates BEV (Bird's Eye View) feature maps and shares them with connected agents. 
The received feature maps are subsequently merged through HPL-ViT which comprises multi-scale graph-attention and bi-directional cross-attention architectures.
Extensive experiments on OPV2V-HPL demonstrate that HPL-ViT consistently attains SOTA (state-of-the-art) results in all experimental scenarios.
It exhibits substantial improvements in scenes with greater heterogeneity.
Besides, HPL-ViT presents impressive generalizability in dynamic scenarios and with varying ego LiDAR, delivering a performance improvement of at least 3.3\% over other fusion methods. 
The main contribution of this paper can be summarized as follows:
\begin{figure}[!t]
\centering
\includegraphics[width=2.2in]{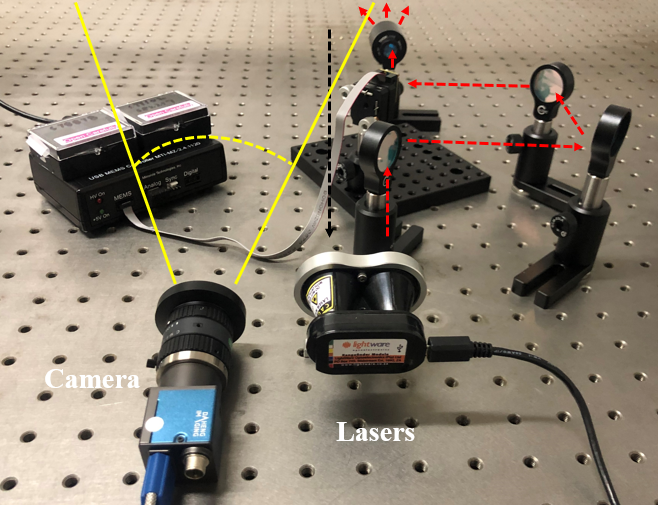}
\caption{Hardware prototype of parallel LiDARs in DAWN platform.}
\label{dawn}
\end{figure}

\begin{itemize}
    \item To the best of our knowledge, this is the pioneering work that explores the heterogeneity of LiDAR systems in V2V. We utilize the combined power of OpenCDA and RLS to construct OPV2V-HPL, which is an enhanced version of OPV2V.
    \item We propose an innovative HPL-ViT framework to improve feature interaction among diverse LiDARs. Our approach incorporates category and frequency encodings for graph attention computations without imposing a significant increase in communication bandwidth. 
    Then bi-directional cross-attention mechanism is introduced to merge features from both category and frequency branches.
    \item Each module in HPL-ViT can be seamlessly integrated into other methods to further improve perception performance. We will soon release our dataset and codes. 
\end{itemize}

\section{Related Work}
\begin{figure*}[!t]
\centering
\includegraphics[width=5.5in]{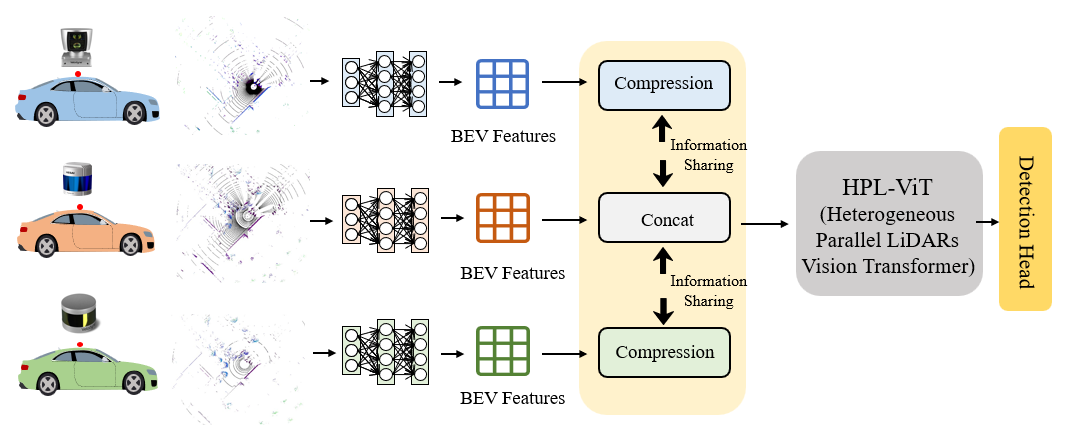}
\caption{The process of cooperative perception with heterogeneous parallel LiDARs in V2V.}
\label{main}
\end{figure*}

\begin{figure}[!t]
\centering
\includegraphics[width=3.4in]{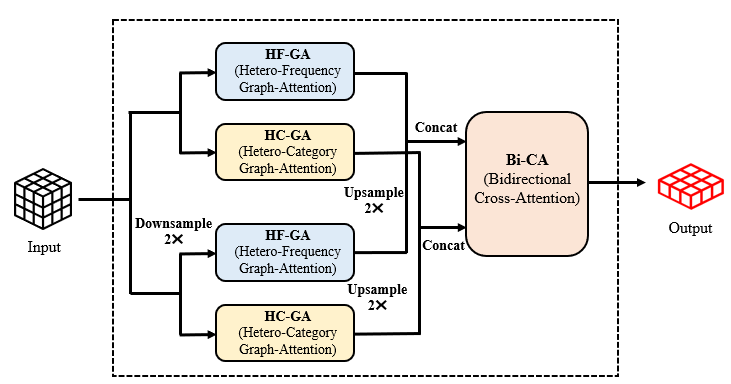}
\caption{HPL-ViT.}
\label{HPL}
\end{figure}
\subsection{Parallel LiDARs}
Parallel LiDARs represent a novel category of intelligent LiDARs founded on parallel intelligence \cite{liu2022parallel}. Parallel intelligence is a pioneering methodological framework introduced by Prof. Fei-Yue Wang \cite{wang2004parallel}, which emphasizes the tight integration of physical and digital realms. 
It has found extensive applications across various domains, including control \cite{wei2020parallel, wang2022dao}, sensing \cite{shen2022parallel, liu2022metasensing}, and autonomous driving \cite{wang2017parallel, liu2020cyber}. 
In the context of parallel sensing, an experimental platform known as DAWN has been constructed \cite{10163892}.
It supports the development of next-generation intelligent sensors such as LiDARs and the light field system \cite{wang2021parallel}. 
\cite{liu2022parallel} proposes the framework of parallel LiDARs which consists of three primary parts: descriptive, predictive, and prescriptive LiDARs. 
Descriptive LiDARs aim to construct complete digital LiDAR systems; predictive LiDARs underscore the significance of computational experiments in cyberspace; while prescriptive LiDARs facilitate real-time interaction between physical and digital LiDARs.
As shown in Figure \ref{dawn}, \cite{10163892} introduces a hardware prototype of parallel LiDAR in DAWN. 
It allows for real-time optimization of perceptual resource allocation assisted by digital LiDARs in high-definition maps. 
\cite{9874864} creates a point cloud dataset of parallel LiDARs under adverse weather, and \cite{liu2022radarverses} discusses the self-maintenance problem. 
In this article, we focus on the robust cooperative perception of heterogeneous parallel LiDARs in autonomous driving.

\subsection{LiDAR-based 3D Object Detection}
LiDAR sensors can provide accurate depth and structural information in autonomous driving which can be used for scene understanding and object identification. 
According to different data representations, the LiDAR-based 3D object detection method can be categorized into four types \cite{mao20223d}. 
\begin{itemize}
\item Point-based methods are specifically designed for the unique structural properties of point clouds which directly extract features at the individual point level. 
PointNet \cite{qi2017pointnet, qi2017pointnet++} stands out as a pioneering work in point-based methods that utilizes MLP to efficiently capture point-wise features and a lot of subsequent works attempt to optimize feature extraction using graph operators \cite{shi2020point} or Transformer architecture \cite{pan20213d}. 
\item Grid-based methods firstly partition point clouds into regular 2D or 3D voxels and extract features using the standard convolution networks in 2D vision tasks. 
VoxelNet \cite{zhou2018voxelnet} is the first work to generate features using 3D CNN and SECOND \cite{yan2018second} proposes a novel sparse convolution to efficiently reduce computation. On the other hand, PointPillars \cite{lang2019pointpillars} projects point clouds into the BEV perspective and adopts 2D CNN for data processing, which offers significant advantages in terms of processing speed. 
\item Point-voxel-based methods use a hybrid architecture to extract features at both point and voxel levels. PV-RCNN \cite{shi2020pv} is a typical model that can learn features from different data representations at each stage. 
\item Range-based methods convert point clouds into an image-like format which can be regarded as a sparse depth map recording range information \cite{meyer2019lasernet}. Advanced 2D object detection approaches can be directly used for feature extraction, providing the potential convenience for multimodal data fusion. 
\end{itemize}

For the sake of real-time performance, we choose PointPillars for intermediate feature extraction in our study. 

\subsection{Cooperative Perception}
Cooperative perception aims to utilize V2V or V2I (Vehicle-to-Infrastructure) technologies to enhance the vehicle's perceptual performance. 
Recent public dataset releases have significantly advanced this field, such as OPV2V \cite{xu2022opv2v} and V2XSet \cite{xu2022v2x} acquired in CARLA, as well as V2V4Real \cite{xu2023v2v4real} and DAIR-V2X \cite{yu2022dair} captured in physical settings.

Current research on V2V focuses on LiDAR-based 3D object detection and it can be divided into three main approaches: early, intermediate, and late fusion. 
Early fusion directly transmits raw point cloud data to the ego car \cite{gao2018object}, while late fusion only sends the generated 3D bounding boxes \cite{song2023cooperative}.
Intermediate fusion \cite{xu2022v2x, lei2022latency, hu2022where2comm, 10077757, xu2022opv2v} extracts neural features at first and then shares them for fusion, achieving a balance between bandwidth and model accuracy. 
V2VNet \cite{wang2020v2vnet} is a pioneering work that proposes a spatial-aware mechanism to share compressed feature maps. 
DiscoNet \cite{li2021learning} adopts a graph-based structure to fuse features, and \cite{xu2022opv2v} suggests a novel single-head self-attention fusion method. 
MPDA \cite{xu2023bridging} is an interesting work that investigates the impact of heterogeneous models on perceptual performance. A learnable resizer module is applied to align feature maps of varying sizes and a domain classifier is used for domain invariant feature extraction. 
\cite{xiang2023hm} also considers the difference between LiDARs and cameras in cooperative perception. 
In contrast to previous works, we focus on the issue of heterogeneous LiDARs in V2V and propose a novel HPL-ViT framework to improve feature fusion performance.

\section{Methodology}
In this work, we address a more practical scenario of V2V in autonomous driving, wherein vehicles are equipped with diverse LiDAR sensors operating at varying frequencies.
Each vehicle can communicate with its surroundings, and our primary focus lies in LiDAR-based 3D object detection. 
As illustrated in Figure 3, the overall framework consists of four main parts: feature extraction, data compression and sharing, HPL-ViT for fusion, and a detection head.

\subsection{Main Architecture} 

\subsubsection{Feature Extraction}
To optimize inference speed, we integrate the grid-based PointPillar for extracting intermediate features from the raw point cloud. 
It initially partitions the point cloud into individual pillars and generates a pseudo image in the BEV perspective.
Then a multi-scale CNN backbone is employed to extract BEV feature maps $F(i) \in R^{H \times W \times C} $. 
$i$ represents the agent index, while $H$, $W$, and $C$ correspond to the height, width, and channels, respectively. 

\subsubsection{Data Compression and Sharing}
To reduce data transmission bandwidth, we use 1$\times$1 convolution kernels to compress the feature map into $F(i)^{'} \in R^{H \times W \times C_0} (C_0 < C)$. 
The compressed feature maps are subsequently transmitted to the ego vehicle for intermediate feature fusion. 
In addition, each agent is required to share information regarding the category and operating frequency of their installed LiDARs.
Given our focus on the heterogeneity of LiDARs in this study, we do not consider factors such as position errors and communication delays during data transmission.

\subsubsection{HPL-ViT}
HPL-ViT is a novel vision transformer architecture that utilizes concatenated feature maps as its input. Figure \ref{HPL} provides an overview of the HPL-ViT framework. 
Our approach begins by applying multi-scale heterogeneous graph-attention mechanisms to enhance feature interactions in both the category and frequency domains.
After aligning the features across multiple scales, we feed them into a bidirectional cross-attention module for fusion.
Although HPL-ViT incorporates multi-scale computations, it retains consistent dimensions for input and output feature maps, preventing the loss of fine-grained details caused by downsampling.

\subsubsection{Detection Head}
We utilize two 1$\times$1 convolution layers as the detection head, with one responsible for bounding box regression and the other for classification. 

\subsection{HPL-ViT}

\subsubsection{Multi-scale Heterogeneous Graph-attention}

\begin{figure}[!t]
\centering
\includegraphics[width=3.2in]{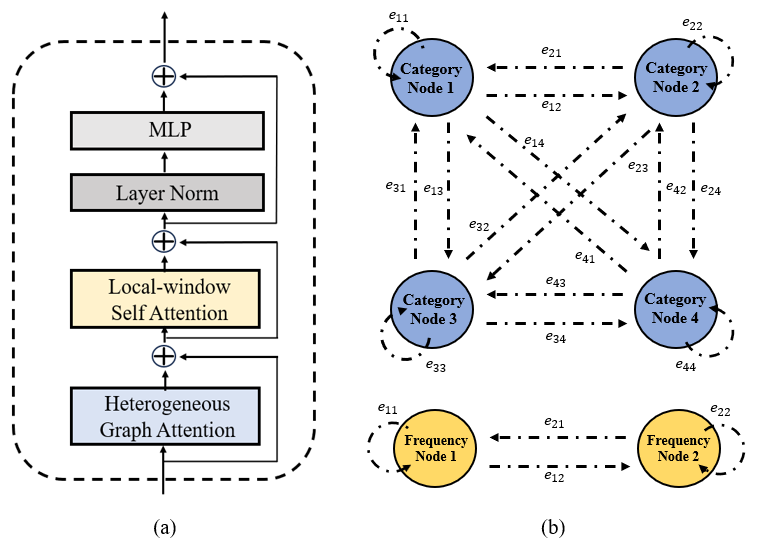}
\caption{(a). A heterogeneous graph-attention block. (b). The constructed heterogeneous graphs.}
\label{H}
\end{figure}

Intermediate features extracted from different LiDARs exhibit distinctive characteristics. 
To address this challenge, we categorize the LiDAR's heterogeneity into two types: category and frequency, and then propose multi-scale heterogeneous graph-attention to capture domain-specific information.
Figure \ref{H}(a) illustrates a complete heterogeneous graph-attention block which comprises a graph-attention and a local self-attention layer. 
The graph-attention layer enhances feature interactions at the same spatial location across different feature maps. 
Additionally, we introduce multi-scale feature computation to extract more comprehensive semantic information at different granularity levels.

Figure \ref{H}(b) illustrates the heterogeneous graphs with each node representing an individual vehicle. 
We take the category heterogeneous graph as an example to explain its details.
Each vehicle node $i$ stores the compressed BEV feature map $F(i)^{'}$ and its installed LiDAR category $C(i)$. 
$E_{ij}$ represents the edge connecting nodes $i$ and $j$. 
Although we use a multi-head attention mechanism, we only present a single-head attention scenario in the description for clarity.
Multi-head attention can be easily achieved by concatenating features extracted from multiple heads. 
Firstly, we utilize liner layers to generate the query $Q(i)$, key $K(i)$, and value $V(i)$ vectors from $F(i)^{'}$. It should be noted that each $C(i)$ corresponds to a unique set of linear layers. 
Furthermore, we define two sets of learnable parameters, $W_{e_{ij}}^{Att}$ and $W_{e_{ij}}^{v}$, which are used for weighting attention maps and values: 


\begin{equation}
Att_x(i, j)=softmax(Q_x(i)W_{e_{ij}}^{Att}K_x(j)), 
\end{equation}
\begin{equation}
Msg_x(i, j) = V_x(j)W_{e_{ij}}^{v}, 
\end{equation}

$x\in R^{2}$ denotes the spatial location in feature maps.
Following that, we finalize the feature update for node $i$:

\begin{equation}
G_x(i) = \sum_{j \in N(i)} Att_x(i, j) Msg_x(i, j),
\end{equation}

$N(i)$ includes all other vehicle nodes engaged in data interaction with node $i$.
We then incorporate a local self-attention layer to promote feature interactions within each node. 
To acquire multi-scale features, we downsample the input features and align them using inverse convolution.
We independently concatenate output features from the frequency and category branches to generate $F_c(i)$ and $F_f(i)$.

\subsubsection{Bidirectional Cross-Attention}

\begin{figure}[!t]
\centering
\includegraphics[width=3in]{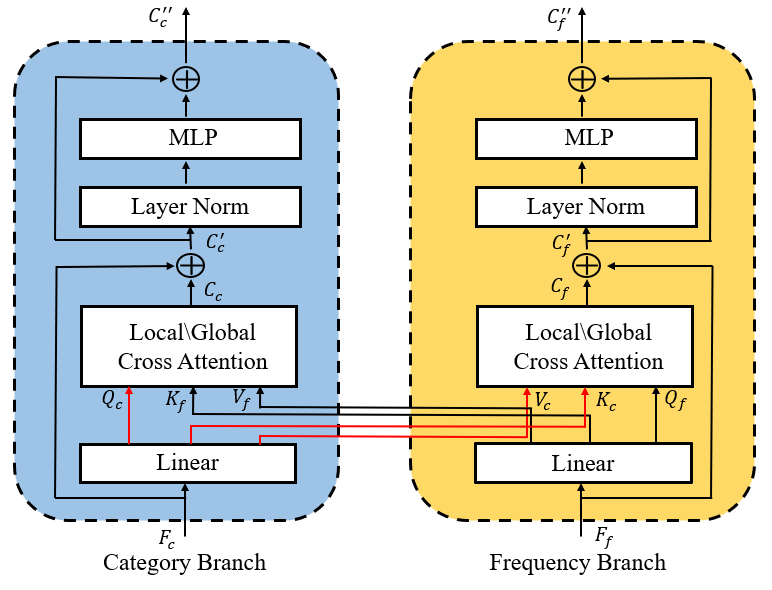}
\caption{A bidirectional cross-attention module.}
\label{B}
\end{figure}

A novel bidirectional architecture stacked with multiple local and global cross-attention blocks is proposed to integrate features from different branches. 
Figure \ref{B} presents the structure of a single cross-attention module.

We begin by aggregating multi-vehicle features into $F_c$ and $F_f \in R^{N \times H \times W \times C}$, where N represents the count of connected vehicles. 
These tensors are then partitioned into $(\frac{H}{m}, \frac{W}{m}, N \times m^2, C)$, where each window includes $Nm^2$ tokens for feature interaction. $m$ denotes the window size. 
The cross-attention module adopts distinct linear layers to generate queries, keys, and values for each branch. 
The query in the current branch is designed to access the key and value components of another branch. 
Following this, the fused feature is processed through an MLP layer with skip connections, ensuring robust gradient backpropagation:

\begin{equation}
C_a=softmax(\frac{Q_a K_b}{\sqrt d_c}) V_b,
\end{equation}
\begin{equation}
C_a^{'} = C_a + F_a,
\end{equation}
\begin{equation}
C_a^{"} = MLP(LN(C_a^{'})) + C_a^{'},
\end{equation}

The pairs $(a, b)$ corresponds to either $(category, frequency)$ or $(frequency, category)$, with $d_c$ representing the feature length. 
$Q$, $K$, and $V$ with subscripts denote the query, key, and value for their respective branches. 
To enhance the receptive field, we adopt an axis-swapping strategy \cite{tu2022maxvit, xu2022cobevt} for extracting global semantic information without introducing additional complexity. 
Local attention efficiently captures object-level details, while global attention uses discrete sampling to aggregate scene-level global information.
We also utilize multi-head attention in our actual implementation.

\section{Experiments}

\subsection{Dataset}
Our experiments utilize OPV2V-HPL, which is an enhanced iteration of OPV2V. 
OPV2V \cite{xu2022opv2v} is the first large-scale V2V dataset collected in CARLA, comprising a total of 11,464 data frames. 
It's split into training, validation, and test sets with 6,764, 1,981, and 2,719 frames, respectively. 
Each data frame incorporates a combination of point cloud and image data, all of which are captured using CARLA's default sensor models.
RLS \cite{10161027} is a high-fidelity LiDAR model library with consistent parameters of physical LiDARs.
We select four different mechanical LiDARs from RLS and replay OPV2V scenarios in OpenCDA for data collection. 
These LiDARs include Hesai Panda64, Velodyne HDL64, VLP32, and VLP16 models, each characterized by varying beam counts and pitch angle distributions. 
It allows for the accurate reconstruction of complex LiDAR interactions in autonomous driving.
Additionally, we also take into account two different operating frequencies: 10 Hz and 20 Hz.

\subsection{Experimental Setup}

\subsubsection{Evaluation Metrics}
We use 3D object detection accuracy as the metric to evaluate different fusion methods. 
To be more specific, we compute the AP (Average Precision) at IoU (Intersection-over-Union) thresholds of 0.5 and 0.7.
In our experiments, we define the detection range as follows: x$\in[-140.8, 140.8]$, and y$\in [-38.4, 38.4]$.

\subsubsection{Implementation Details}
PointPillars is utilized to generate intermediate feature maps, which are subsequently fed into HPL-ViT. 
We also choose six other methods for comparative analysis, including no fusion baseline, OPV2V \cite{xu2022opv2v}, V2VNet \cite{wang2020v2vnet}, F-Cooper \cite{chen2019f}, CoBEVT \cite{xu2022cobevt}, and late fusion.
We train all the models for 30 epochs using 8 Nvidia V100 GPUs and adopt the Adam optimizer. 
The initial learning rate is configured at 0.001, and we employ the cosine annealing with a warm-up strategy to dynamically adjust the learning rate.
Our loss function comprises two primary components: a smooth L1 regression loss with a coefficient of 1, and a focal classification loss with a coefficient of 2.


\subsubsection{Experimental Scenarios}
We establish three distinct scenarios for evaluation under the assumption that all ego vehicles are equipped with 20Hz Panda64 LiDARs.

\begin{itemize}
    \item \textbf{Normal scenario}: In the normal scenario, all other agents employ a 20Hz Panda64 LiDAR, ensuring consistency with the LiDAR system of the ego vehicle.
    \item \textbf{Hetero scenario 1}: 
    Vehicles in hetero scenario 1 are outfitted with different LiDARs which are chosen randomly from the four LiDAR devices stated above. 
    Despite the diversity in LiDAR types, they all operate at a uniform frequency of 20 Hz.
    \item \textbf{Hetero scenario 2}: Hetero scenario 2 represents a more intricate and realistic setting, considering both category and frequency heterogeneity.
\end{itemize}

\subsection{Quantitative Evaluation}

\begin{figure}[!t]
\centering
\includegraphics[width=3.5in]{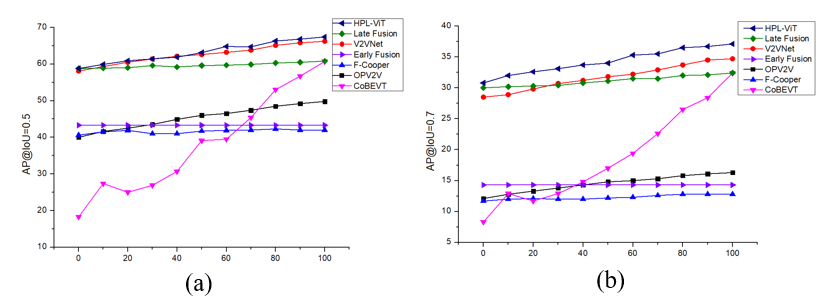}
\caption{Generalization performance in dynamic scenarios.}
\label{hetero1}
\end{figure}

\begin{figure}[!t]
\centering
\includegraphics[width=3.5in]{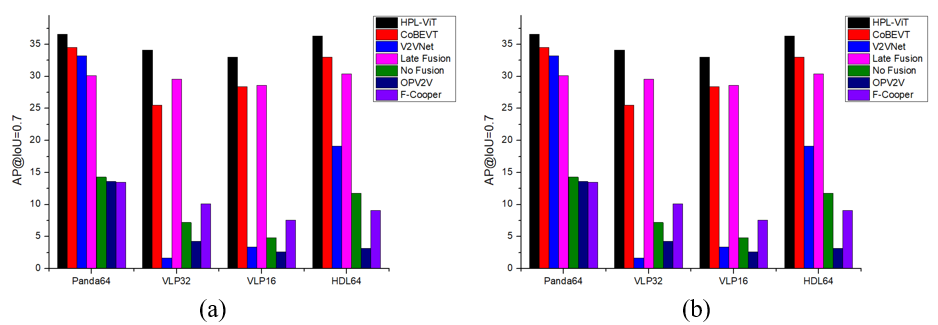}
\caption{Generalization performance with varying ego LiDARs.}
\label{hetero2}
\end{figure}

\begin{table*}[htbp]
\renewcommand\arraystretch{1.5}
\caption{3D object detection accuracy in different scenarios.}
\begin{center}
\begin{tabular}{c|c|c|c}
\hline
 & Normal scenario & Hetero scenario 1 & Hetero scenario 2\\ 
Methods & \textbf{AP@0.5 \quad\qquad\qquad AP@0.7} & \textbf{AP@0.5 \quad\qquad\qquad AP@0.7} & \textbf{AP@0.5 \quad \qquad\qquad AP@0.7} \\
\hline
No Fusion & {43.3 \qquad\qquad\qquad 14.3} & {43.3 \qquad\qquad\qquad 14.3} & {43.3 \qquad\qquad\qquad 14.3} \\
Late Fusion & {64.6 \qquad\qquad\qquad 37.3} & {57.5 \qquad\qquad\qquad 29.2} & { 59.2 \qquad\qquad\qquad 30.1} \\
\hline
OPV2V \cite{xu2022opv2v} & {51.1 \qquad\qquad\qquad 18.2} & {48.1 \qquad\qquad\qquad 16.0} & {44.0 \qquad\qquad\qquad 13.6} \\
V2VNet \cite{wang2020v2vnet} & {66.3 \qquad\qquad\qquad 37.3} & {\underline{64.8} \qquad\qquad\qquad 35.3} & {\underline{64.3} \qquad\qquad\qquad 33.2} \\
F-Cooper \cite{chen2019f} & {49.3 \qquad\qquad\qquad 19.3} & {48.3 \qquad\qquad\qquad 15.4} & {46.5 \qquad\qquad\qquad 13.5} \\
CoBEVT \cite{xu2022cobevt} & {\underline{66.7} \qquad\qquad\qquad \underline{37.8}} & {64.4 \qquad\qquad\qquad \underline{35.4}} & {64.0 \qquad\qquad\qquad \underline{34.5}} \\
\hline
HPL-ViT & {\textbf{67.3 (+0.6)} \qquad \textbf{38.3 (+0.5)}} & {\textbf{65.9 (+1.1)} \qquad \textbf{36.4 (+1.0)}} & {\textbf{66.4 (+2.1)} \qquad \textbf{36.6 (+2.1)}} \\
\hline
\end{tabular}
\label{main}
\end{center}
\end{table*}

\begin{figure*}[!t]
\centering
\includegraphics[width=6in]{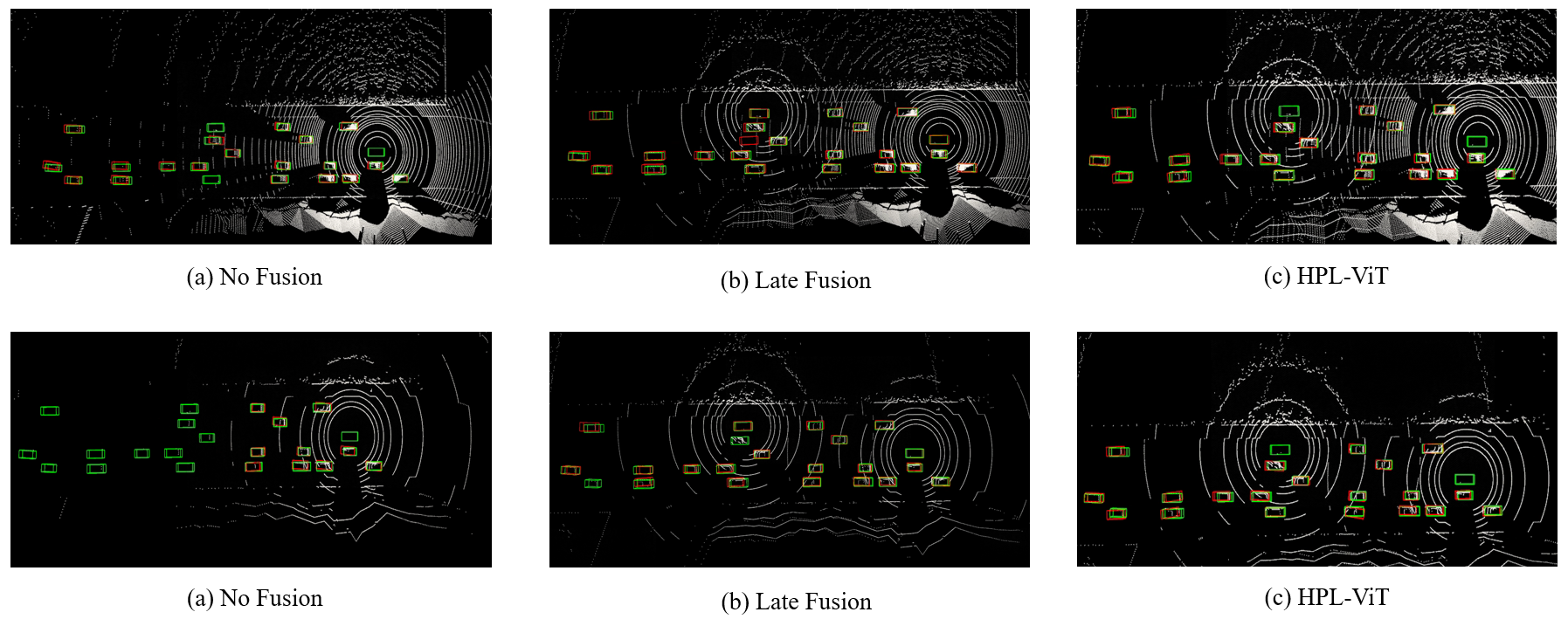}
\caption{3D object detection visualization. \textcolor{green}{Green} and \textcolor{red}{red} bounding boxes represent the \textcolor{green}{ground truth} and \textcolor{red}{predicted}, respectively. 
(a)-(c) are visualization results in Hetero scenario 2 with Panda64 as the ego LiDAR, while (d)-(e) use VLP16 as the ego LiDAR.}
\label{vis}
\end{figure*}

\subsubsection{Main Performance Analysis}

Table \ref{main} presents the perceptual performance of all methods across different settings. 
We observe that nearly all fusion methods have surpassed the performance of no fusion baseline, and HPL-ViT can achieve SOTA results in all scenarios. 
It's worth noting that the introduction of heterogeneous scenarios has had a negative impact on all models, while HPL-ViT exhibits the least degradation. 
As the degree of LiDAR's heterogeneity increases, we find a widening accuracy gap between HPL-ViT and the second-ranked method.
It can reach a significant 2.1\% improvement in hetero scenario 2, which effectively demonstrates its outstanding feature fusion capabilities in complex and challenging environments.


\subsubsection{Generalization Analysis}
To evaluate the generalization capabilities of HPL-ViT, we utilize models trained in hetero scenario 2 and subject them to testing in a broader range of settings. 
First, we fix the ego LiDAR as a Panda64 operating at 20Hz and adjust the ratio of other connected LiDARs to create different scenarios. 
As illustrated in Figure \ref{hetero1}, the horizontal axis signifies the proportion of collaborators equipped with the same Panda64 at 20Hz, and it is evident that HPL-ViT consistently achieves SOTA performance across all configurations.
It's worth noting that late fusion has outperformed most intermediate fusion methods in our experiments.
Additionally, we introduce a change in the type of ego LiDAR to further assess models, with results plotted in Figure \ref{hetero2}. 
We observe that many methods experience a significant decrease in detection accuracy, while HPL-ViT consistently maintains its top-ranking position. 
When deployed on different LiDAR devices, it exhibits a noteworthy accuracy improvement of at least 3.3\%.

\subsubsection{Ablation Study}
We conduct comprehensive ablation studies to evaluate the influence of each component in HPL-ViT.
As depicted in Table \ref{ab}, each component contributes positively to object detection accuracy. 
The absence of HC-GA or HF-GA has a similar adverse effect, while the omission of multi-scale strategy results in the most significant 2.3\% drop in AP@0.7.

\begin{table}[htbp]
\renewcommand\arraystretch{1.5}
\caption{Ablation studies of each component in HPL-ViT.}
\begin{center}
\begin{tabular}{ccccc}
\hline
HC-GA & HF-GA & Bi-CA & Multi-scale & AP@0.5 / AP@0.7 \\ 
\hline
 & \checkmark & \checkmark & \checkmark & 65.2 / 36.2\\
\checkmark &  & \checkmark & \checkmark & 65.2 / 35.7\\
\checkmark & \checkmark &  & \checkmark & 65.4 / 35.4\\
\checkmark & \checkmark & \checkmark &  & 64.1 / 34.1\\
\hline
\checkmark & \checkmark & \checkmark & \checkmark & 66.4 / 36.6\\
\hline
\end{tabular}
\label{ab}
\end{center}
\end{table}




\subsection{Qualitative Evaluation}

In Figure \ref{vis} (a)-(c), we provide visualizations of no fusion, late fusion, and HPL-ViT in hetero scenario 2, respectively.
Then we change the ego LiDAR to a VLP16 operating at 20Hz to present their generalization capabilities, and the testing results are plotted in Figure \ref{vis} (d)-(f). 
It is evident that the no fusion baseline displays numerous missing detections. Although the detection results exhibit noticeable improvement with late fusion, there are still instances of missed detections and false positives. 
In contrast, when compared to the aforementioned methods, HPL-ViT can further enhance detection accuracy by not only eliminating false positives but also attaining more precise bounding box positions.

\section{CONCLUSIONS}

In this paper, we focus on cooperative perception in autonomous driving to facilitate the construction of parallel LiDAR networks. 
This is the first work that investigates the issue of heterogeneous LiDAR sensors in V2V. 
We collect a novel dataset named OPV2V-HPL and present HPL-ViT, which is a vision Transformer architecture for robust feature fusion among heterogeneous parallel LiDARs. 
Extensive experiments have demonstrated that HPL-ViT can attain SOTA performance with strong generalizability in all scenarios. 
In future work, we will extend our research into real-world scenarios and remain committed to the development of a networked parallel LiDAR system.



\bibliographystyle{IEEEtran}      
\bibliography{root.bib}

\end{document}